\documentclass[11pt,a4paper]{article}
\usepackage[hyperref]{acl2018}
\usepackage{times}
\usepackage{latexsym}
\usepackage[justification=centering]{caption}
\usepackage{graphicx}
\usepackage{tabularx}
\usepackage{soul}
\usepackage{algorithm,algorithmic}
\usepackage{textgreek}
\usepackage{multirow}
\usepackage{enumitem}
\usepackage{tabu}
\usepackage{array}
\usepackage{booktabs}
\usepackage{tikz}
\usepackage[normalem]{ulem}
\usepackage[export]{adjustbox}
\usepackage{xcolor}
\setlist{noitemsep}
\usepackage{url}

\aclfinalcopy 

\title{Language Identification of Bengali-English Code-Mixed data using Character \& Phonetic based LSTM Models}

\author{
        Soumil Mandal\textsuperscript{1}, Sourya Dipta Das\textsuperscript{2}, Dipankar Das\textsuperscript{3} \\ \\
       	\textsuperscript{1}Department of CSE, SRM University, Chennai, India\\
        \textsuperscript{2}Department of ETCE,
        Jadavpur University, Kolkata, India\\
        \textsuperscript{3}Department of CSE,
        Jadavpur University, Kolkata, India\\
        \textcolor{black!50}{\{soumil.mandal\textsuperscript{1}, dipta.math\textsuperscript{2}, dipankar.dipnil2005\textsuperscript{3}\}@gmail.com}                
}

\date{}
\begin{document}
\maketitle

\newcommand*\circled[1]{\tikz[baseline=(char.base)]{
            \node[shape=circle,draw,inner sep=1pt] (char) {#1};}}

\begin{abstract}
Language identification of social media text still remains a challenging task due to properties like code-mixing and inconsistent phonetic transliterations. In this paper, we present a supervised learning approach for language identification at the word level of low resource Bengali-English code-mixed data taken from social media. We employ two methods of word encoding, namely character based and root phone based to train our deep LSTM models. Utilizing these two models we created two ensemble models using stacking and threshold technique which gave 91.78\% and 92.35\% accuracies respectively on our testing data. 
\end{abstract}

\section{Introduction}

Exploration of social media and exponential growth of data redirect us to bring various insights on language aspects. For example, a phenomenon known as code-mixing where people from multilingual backgrounds often mix two languages while communicating on social media. As our traditional NLP tools often fail or give poor results on such data \cite{bergsma2012language}, it becomes an important prerequisite to identify language tags with high accuracy. Language identification on such data at both word and sentence level still remains as a challenging task due to a few facts such as inconsistent phonetic transliteration, borrowing of words, spelling errors, intra sentential change of matrix, embedded languages as well as the use of numeric and special characters in words. 
Here, we discuss our system for word level language identification of low resource Bengali-English code-mixed data taken from Twitter, where both are typed in Roman script. Bengali is an Indo-Aryan language of India where 8.10\% \footnote{https://en.wikipedia.org/wiki/Languages\_of\_India} of the total population are the $1^{st}$ language speakers and is also the official language of Bangladesh. The original script in which Bengali is written by locals is the Eastern Nagari Script~\footnote{https://www.omniglot.com/writing/bengali.htm}. Most of the code-mixed data was collected from Twitter~\footnote{https://twitter.com/}. \\
\hspace*{0.5cm}Employing LSTM, two models were developed, one using character based encoding and the other using Bengali root phones. Combining these two models, two ensemble models were prepared using stacking and threshold. Finally, the performance metrics like accuracy and macro-averaged F1-Score along with confusion matrix were used to compare the results. With respect to the related works in Sec~\ref{sec2}, our system managed to score comparable results using a fraction of data.

\section{Related Work}
\label{sec2}
The LID problem in code-mixed texts has already been studied by many researchers in the aspect of social media communication utilizing various methods involving both supervised and unsupervised techniques. \citet{nguyen2013word} experimented with linear chain CRFs but their contextual features are limited to bi-grams of words. \citet{kim2014sociolinguistic} studied the linguistic behavior of bilingual Twitter users from Qatar, Switzerland and Québec, and also acknowledged that code-mixing could not be studied due to the absence of appropriate LID tools. \citet{das2014identifying} introduced code-mixing index to evaluate the level of mixing of the languages and made a system for Bengali-English LID using SVM trained on n-gram with weights, dictionary, minimum edit distance, 7 word window for context and got an F1-Score of 76.37\%. \citet{barman2014code} uses char-n-grams, capitalization, dictionary along with a decision tree trained on word length to train a SVM model for code-mixed LID. They also use CRF for context capturing. \citet{jhamtani2014word} combined to classifiers into an ensemble model for Hindi-English code-mixed LID. The first classifier used modified edit distance, word frequency and character n-grams as features. The second classifier used output of the first classifier for the current word, along with language tag and POS tag of neighboring to give the final tag. \citet{piergallini2016word} used n-grams along with capitalization as features to train a linear model for Swahili-English word level LID and got an accuracy of 96.5\%. \citet{jaech2016neural} employed a hierarchical neural model, where a CNN was trained on char2vec and bi-LSTM on sentence level word embeddings to create a system which got an F1-Score of 93.1\% on Spanish-English data. \citet{rijhwani2017estimating} demonstrates a generalized unsupervised model based on HMM for code-mixed LID for an arbitrarily large set of languages. \citet{singer2003acoustic} used phones and acoustic properties for language identification in speech data.  To the best of our knowledge, none of the works have used character or phonetic encoding in an LSTM architecture for textual language identification task.      

\section{Data Sets}
We collected transliterated Bengali words in Roman and English words from  ICON 16~\footnote{http://ltrc.iiit.ac.in/icon2016/}, ICON 17~\footnote{https://ltrc.iiit.ac.in/icon2017/} contests and code-mixed data used in \citet{mandal2018analyzing}. Some additional English words were collected from online~\footnote{https://github.com/first20hours/google-10000-english} resources. The datasets had no intersection. All words were converted to lowercase, and words having length less than 3 or any special or numeric characters (e.g. \textit{ri8}, \textit{2toh}) were removed. Also, words with $>$ 2 consecutive identical characters was normalized to 2 consecutive identical characters. Words which had both EN and BN components, for e.g. \textit{journey(ta)}, which is an EN word with BN suffix were discarded as well. The distribution for training, development and testing was 6632x2, 300x2 and 700x2, respectively. It should be noted that the amount of data is comparatively less as Bengali falls in the low resource language category when dealing with data from social media.

\section{Support Vector Machine Model}
In order to start with (baseline model), we decided to make a supervised model using Support Vector Machine (SVM) with bag-of-words (BOW) principle. The training and development set were merged in order to create the training data in this case. Then character level bigrams, trigrams and quadgrams were extracted and SVM with linear kernel implemented in the scikit-learn~\footnote{http://scikit-learn.org/stable/} package was employed for training purpose. This system achieved an accuracy of 83.64\% on the test data.    

\section{Neural Model}
We moved towards implementing neural models to achieve better accuracy. Two encoding methods for neural model were tried, namely character based and phone based. The encoding methods we used are described in detail below. 

\subsection{Character Encoding}

We decided to make a neural model based on character encoding for acquiring patterns at an elementary level. Similar encoding methods have been tested for character level NMT before, used in a seq2seq architecture \cite{lee2016fully}. In order to build the encoder, a dictionary based on character-index was made where index is the $n^{th}$ English alphabet. The algorithm fetches each character at a time from the word and replaces it with the respective index, for example, the encoding of \textit{good} is represented as [7,15,15,4], for \textit{bad} it will be [2,1,4].  

\subsection{Phonetic Encoding}

It is evident that different languages have different phonetic properties~\footnote{https://en.wikipedia.org/wiki/Phonetics}, especially when they have distant origins. In order to exploit this, we developed a language identification neural model based on phonetic encoding. One of our major aims was to obtain such encodings in terms of Bengali root phones. Thus, a phonetic library that contains two parts, namely root phones and similar phone groups was prepared by the authors in 3 steps. First, character level bigrams and trigrams were collected from the Bengali words present in the training set and were sorted in non-increasing order of frequency. We did not extend to quadgrams as we could see by observing the data set that the users generally tend to use maximum of 3 characters to phonetically represent the corresponding Bengali character. Then, from the code chart of Bengali unicode characters~\footnote{https://unicode.org/charts/PDF/Unicode-10.0/}, the pronunciation of each of the Bengali characters written in Roman script was gathered and finally, a root phone list (RP) was prepared. By referring to the (2,3)-gram frequency table, we noticed that some of the transliterations collected from the unicode chart are rarely used in real life for phonetic typing. Thus, such instances were discarded from the list (e.g. \textit{nga}, \textit{nya}, \textit{ddha}). Next, with respect to the root phones, all the corresponding (phonetically similar) bigrams and trigrams with high frequency were gathered and formed into groups, i.e similar phone groups (SPG). Though some of the bigrams and trigrams had quite high frequency, yet they were not included in SPG as they didn't correspond to any of the root phones, example (e.g. \textit{bhi}). The phonetic library was finally checked and verified by a linguist. Complete model consisting of RP and SPG is shown in Table~\ref{table2}. In each of the phonetic groups, we had set the first phone as the respective root phone for ease of searching and retrieval of root index. 

\begin{figure}[h]
\includegraphics[scale=0.4, center]{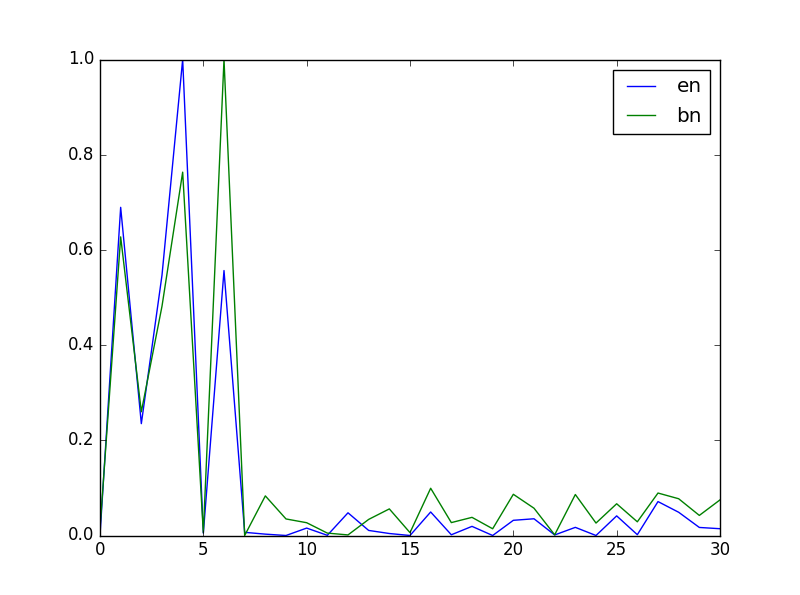}
\caption{Normalized root phones frequency.}
\label{fig1}
\end{figure}

\begin{table}[H]
\centering
\begin{tabular}{|c|l|l|l|}
\hline
\multicolumn{4}{|c|}{\textbf{Root Phones}} \\ \hline
\multicolumn{4}{|c|}{\begin{tabular}[c]{@{}c@{}}aa, i, u, r, e, ai, o, au, ka, kha, ga, gha, ca, cha,\\ ja, jha, ta, tha, da, dha, na, pa, pha, ba, bha, ma,\\ ya, ra, la, sa, ha\end{tabular}} \\ \hline
\multicolumn{4}{|c|}{\textbf{Phonetic Groups}} \\ \hline
\multicolumn{4}{|c|}{(aa, a), (i, ee), (u, w), (r, ri), (e), (ai, oi), (o, oo)} \\ \hline
\multicolumn{4}{|c|}{(au, ou, ow), (ka, k), (kha, kh), (ga, g), (gha, gh)} \\ \hline
\multicolumn{4}{|c|}{(ca, c), (cha, ch), (sa, s, sh), (jha, jh), (bha, bh, v)} \\ \hline
\multicolumn{4}{|c|}{(ta, t), (tha, th), (da, d), (dha, dh), (na, n), (pa, p)} \\ \hline
\multicolumn{4}{|c|}{(pha, ph, f), (ba, b), (ma, m), (ya, y), (ra, rh)} \\ \hline
\multicolumn{4}{|c|}{(la, l), (ja, j, z), (ha, h)} \\ \hline
\end{tabular}
\caption{Phonetic library.}
\label{table2}
\end{table}
\noindent The routine for encoding is based on variable length sliding window technique, where n$\leftarrow$[3,1].  The algorithm in each of iteration takes n characters, with n starting as 3. If the character cluster appear in phonetic groups, we append the root phone index, else we reduces n by 1 and repeat the same process till n = 1. If the n-gram is not found in phonetic groups, 35 (as number of root phones is 31) is appended. From the next iteration, the window starts from n + 1, where n denotes the ending point of the window in the last iteration. The algorithm stops when window traverses the whole word. Pseudo-code given below.
\\ \\
\textit{while} index $<$ len(word)\\
\hspace*{0.5cm}\textit{if} word[index:index+j](j$\rightarrow$3,2,1) \textit{in} SPG\\
\hspace*{0.5cm}\hspace*{0.5cm}enc.append(root\_phone\_index), index += j\\
\hspace*{0.5cm}\textit{else} enc.append(35), index += 1\\
\textit{return} enc\\
\\
On encoding the words in the combined dataset using the above algorithm, the number of times 35 occurred (i.e n-gram absent in the phonelib is seen) was found to be 82 and 1 for EN and BN data respectively. Some examples of word encodings are, \textit{khabar} meaning 'food' will be [10,24,4], which is essentially \textit{kha-ba-r}. Also \textit{khbr}, which is the same word but is less seen in terms of phonetic typing, produces identical encoding. Though this transforming ability performs quite well on many of such instances, it fails on some instances as well like the word \textit{korchi} produces [9,7,4,14,2] while \textit{krci} results in [9,4,13,2] even though they are the phonetic transliterations of the same word, where the latter is less accurately typed. The normalized root phones frequency graph is shown in Fig~\ref{fig1}. There, y-axis denotes the normalized frequency while the x-axis denotes the root phones (same order as in Table~\ref{table2}). From the table we can see 3 spikes in the starting, which are essentially the frequency of the phones \textit{i},\textit{e},\textit{o}. Thus we can draw the inference, that these 3 RP are quite essential while transliterating BN in Roman. From the rest of the graph, we can see a clear distinction between the blue and green line, implying how EN and BN is different in terms of the root phones.

\section{Training}
We adopted the Long Short Term Memory version of the RNN architecture~\footnote{https://keras.io/} to train both of our models. We specifically chose this because it outperforms almost all the other models in sequence learning \cite{greff2017lstm}. Two LSTM models were built, one for each type of encoding. We set the same parameters for both, with sigmoid activation function and adam optimizer  whereas the loss function used was binary cross-entropy, number of epochs was set to 500 and batch size was set to 1658. All the other parameters were kept at default. The significant difference between the two models was the architecture. For character encoding, the layer sizes were 15-35-25-1 and for phonetic it was 15-15-40-1. The numerics indicate here that the input and the output layer has 15 and 1 node(s) respectively, and the other two are the number of nodes in two hidden layers. The encodings were padded to length 15 in order to make them uniform and was finally converted into one-hot vectors. The target label for Bengali and English was set to 0 and 1, respectively, thus the output of the trained models was fuzzy values between them (this can be visualized clearly in the scatter plots given in the next section).

\section{Evaluation}
For both the character and phonetic models, first a simple roundup logic was used for testing on the development data (dev\_round). As from the fuzzy value scatter plots we could see that theres a scope of improvement, we decided to use brute force for finding a threshold on the development data (dev\_thresh) in order to tune the trained model. Finally using this threshold, we evaluated the model on our testing data (test\_thresh). The details of both the models along with the ensemble model is given below. 

\subsection{Character Model}
\begin{figure}[h]
\includegraphics[scale=0.4, center]{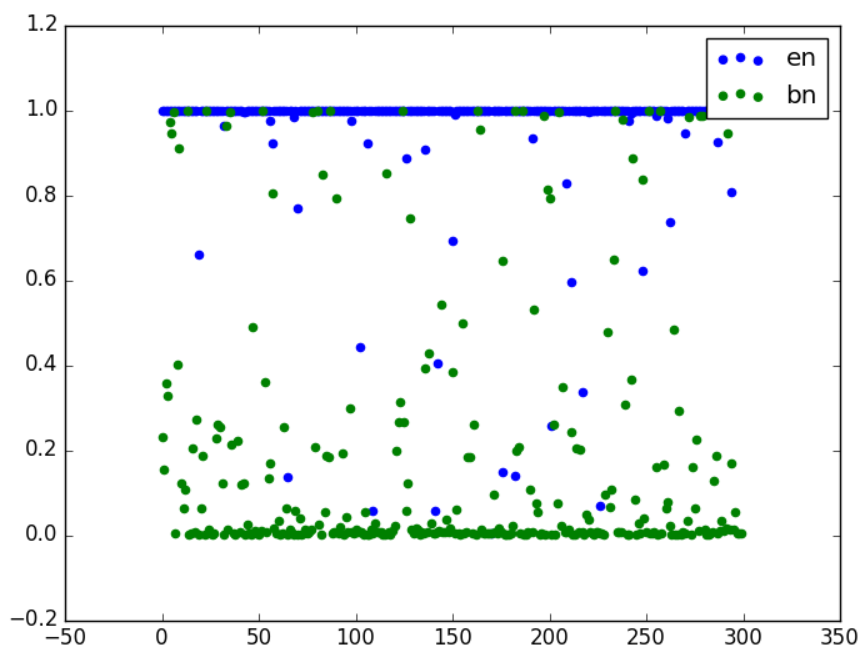}
\caption{Scatter plot of character model fuzzy values on development data.}
\label{fig2}
\end{figure}

The results of the character model are shown in Table~\ref{table4} and the fuzzy outputs scatter plot on the development data is shown in Fig~\ref{fig2}. Simple round up on the fuzzy output gave an accuracy of 91.50\%. Using brute force, the threshold where the accuracy is peaking on the development data was found to be at $\theta$ $\leq$ 0.92. This showed a slight improvement in accuracy by 0.66\%. Finally, on test data, this tuned model gave an accuracy of 91.71\%. 

\begin{table}[H]
\centering
\begin{tabular}{|c|c|c|c|c|}
\hline
\textbf{Model} & \textbf{Acc} & \textbf{Prec} & \textbf{Rec} & \textbf{F1} \\ \hline
dev\_round & 91.50 & 87.84 & 96.33 & 91.48 \\ \hline
dev\_thresh & 92.16 & 91.47 & 93.00 & 92.16 \\ \hline
test\_thresh & 91.71  & 91.59  & 91.85  & 91.71  \\ \hline
\end{tabular}
\caption{Character model results (in \%).}
\label{table4}
\end{table}

\subsection{Phonetic Model}
\begin{figure}[h]
\includegraphics[scale=0.4, center]{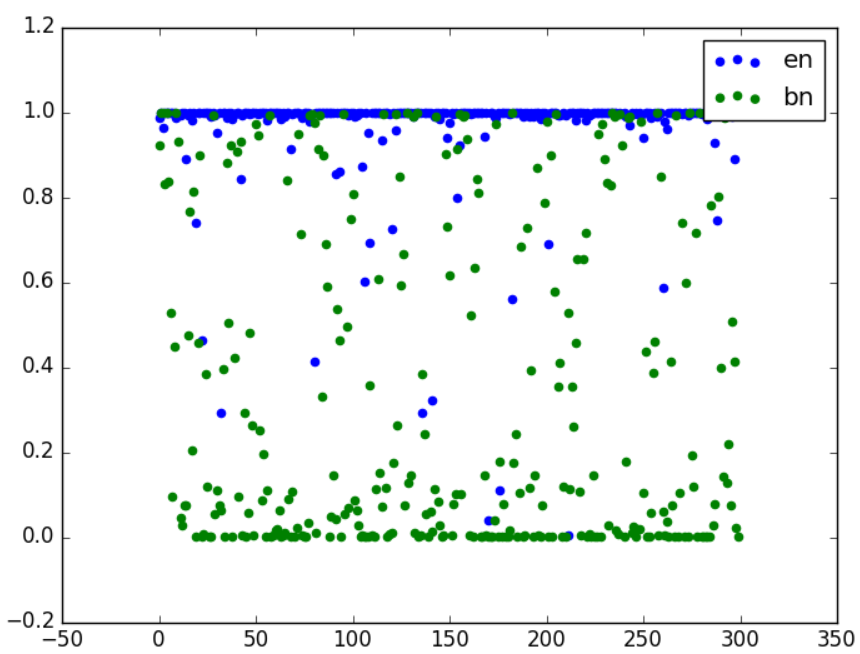}
\caption{Scatter plot of phonetic model fuzzy values on development data.}
\label{fig3}
\end{figure}

Results of the phonetic model are shown in Table~\ref{table5}. On simple roundup on development data, the accuracy was not very good (82.5\%), but using a similar threshold technique as used in the character model, the accuracy improved significantly (6.16\%) and reached 88.66\%. The threshold in this case was calculated to be $\theta$ $\leq$ 0.95. The scatter plot of the fuzzy values is shown in Fig~\ref{fig3}. Using threshold, the accuracy achieved on the test data was 90.42\% and a precision of 91.74\% which is slightly better than the character model.

\begin{table}[H]
\centering
\begin{tabular}{|c|c|c|c|c|}
\hline
\textbf{Model} & \textbf{Acc} & \textbf{Prec} & \textbf{Rec} & \textbf{F1} \\ \hline
dev\_round & 82.50 & 75.06 & 97.33 & 82.10 \\ \hline
dev\_thresh & \dashuline{88.66} & 87.66 & 90.00 & 88.66 \\ \hline
test\_thresh & 90.42  & 91.74  & 88.85  & 90.42  \\ \hline
\end{tabular}
\caption{Phonetic model results (in \%).}
\label{table5}
\end{table}

\subsection{Ensemble Model}
Two ensemble methods were tried, using stacking and thresholding. For the former, logistic regression was used as the combiner algorithm and was trained on the fuzzy values given by the two models on the development data. For threshold, mean of the two fuzzy values was taken and by using a similar brute force technique as used in char and phonetic model, the threshold was found to be $\theta$ $\leq$ 0.9. The architecture is shown below in Fig~\ref{fig4}. 

\begin{figure}[h]
\includegraphics[scale=0.25, center]{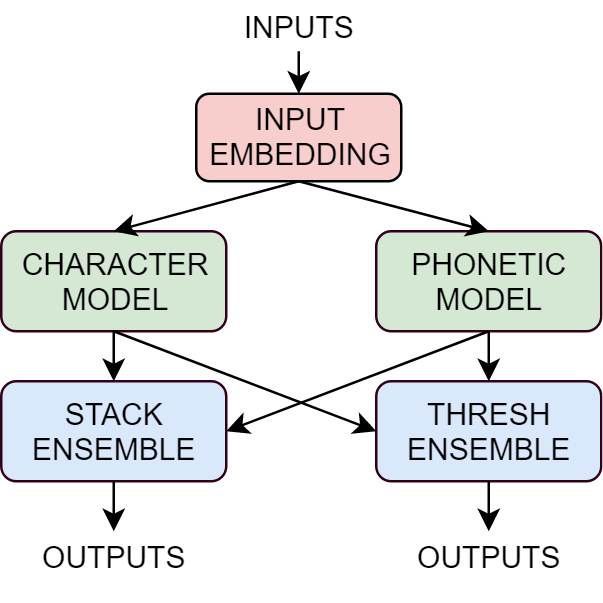}
\caption{Ensemble architecture.}
\label{fig4}
\end{figure}

\begin{figure}[h]
\includegraphics[scale=0.4, center]{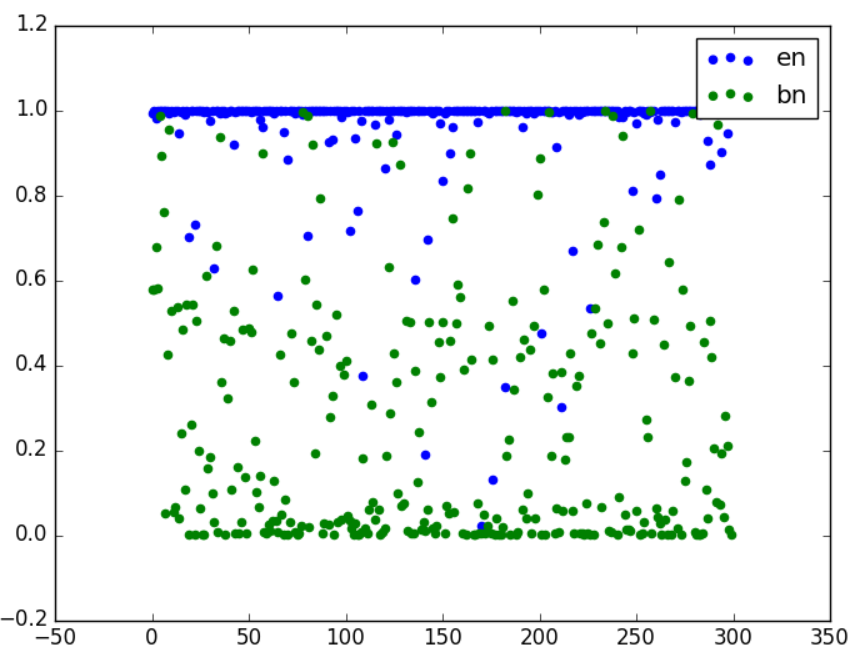}
\caption{Scatter plot of ensemble model fuzzy values on development data.}
\label{fig5}
\end{figure}

\noindent The results compared with the baseline model is shown in Table~\ref{table6} and the scatter plot of means is shown in Fig~\ref{fig5}. 

\begin{table}[H]
\centering
\begin{tabular}{|c|c|c|c|c|}
\hline
\textbf{Model} & \textbf{Acc} & \textbf{Prec} & \textbf{Rec} & \textbf{F1} \\ \hline
svm\_baseline & 83.64 & 83.88 & 83.28 & 83.64 \\ \hline
ensem\_stack & 91.78  & 89.58  & 94.57  & 91.77  \\ \hline
ensem\_thresh & \bf{92.35}  & 94.99  & 89.42  & 92.35  \\ \hline
\end{tabular}
\caption{Ensemble model results (in \%).}
\label{table6}
\end{table}

\noindent The stacking method showed slight improvement and the accuracy achieved was 91.78\%. On the other hand, accuracy improved by 0.57\% using threshold technique. Also, we can see the improvement using the neural networks is quite noticeable, i.e an increase in accuracy by 8.71\% from the SVM n-gram based model.

\subsection{Error Analysis}
We started our error analysis by preparing confusion matrices (CM) of the four models. The values observed on test data are shown in Table~\ref{table3}. Here, BN correctly predicted was considered as TP and EN was considered as TN.
\newcolumntype{?}{!{\vrule width 1.3pt}}
\begin{table}[H]
\centering
\begin{tabular}{|c|c|c?c|c|c|}
\hline
\multicolumn{6}{|c|}{\textbf{Confusion Matrices}} \\ \hline
\circled{1} & \textit{BN} & \textit{EN} & \circled{2} & \textit{BN} & \textit{EN} \\ \hline
BN & 641 & 59 & BN & 644 & 56 \\ \hline
EN & 57 & 643 & EN & 78 & 622 \\ \specialrule{1.2pt}{1pt}{1pt}
\circled{3} & \textit{BN} & \textit{EN} & \circled{4} & \textit{BN} & \textit{EN} \\ \hline
BN & 623 & 77 & BN & 667 & 33 \\ \hline
EN & 38 & 662 & EN & 74 & 626 \\ \hline
\end{tabular}
\caption{Confusion matrices. (1-character, 2-phonetic, 3-ensem\_stack, 4-ensem\_thresh)}
\label{table3}
\end{table}
\noindent From CM 1 given in Table~\ref{table3}, we found that the character model is quite balanced in terms of biasness as the values of TP and TN are quite close. Unlike the character model, we can see from CM 2 (phonetic model) that TP is much more than TN, thus the precision is higher by 0.15\% compared to the character model. 
From CM 3, we observed that TN is much higher compared to TP (by 39) and thus, it can be inferred that the logistic regression model is biased towards EN. On the contrary, from CM 4, we can claim that TP is much higher compared to TN (by 41). This is due to the fact that the regression range of the fuzzy outputs from the models for BN is much higher compared to that of EN, and thus calculating threshold keeping accuracy as the metric favors TP.

\section{Observations}
One of the primary drawbacks of our system is that it is based on word level and not sentence, thus, it fails in capturing context information. This problem is quite evident when words with similar spellings but belonging to different languages are assigned with same tag. E.g, in our test data, the word "\textit{choke}" was present in both EN and BN (meaning \textit{eyes}) data but was always tagged as BN. The system is not designed to handle words with numeric or special characters (e.g. \textit{ri8} instead of \textit{right}), and also elongated words (e.g. \textit{goood} instead of \textit{good}) though a simple normalization prior to feeding would solve the issue. 

\section{Conclusion \& Future Work}
In the present work, we have build a system for word level LID. A new method for word encoding has been introduced using root phones. Employing character and phonetic encoding methods, we designed two deep LSTM models. Finally, two ensemble models were designed using stacking and threshold techniques. The stacking method achieved an accuracy of 91.78\% while on the other hand the threshold model obtained the best accuracy with a score of 92.35\%. Considering that our training data was quite low $\approx$13k, the architecture seems to be quite promising. In future, our aim is to gather more data from social media. This would not only help in the generic improvement of accuracy, but the n-gram statistics may also be useful to enrich and fine tune our phonetic library even more, thus improving the encoding quality. Also, we would like to test a similar methodology on other language pairs as well to evaluate its applicability.

\bibliography{acl2018}
\bibliographystyle{acl_natbib}

\end{document}